\documentclass[12pt]{article}

\usepackage[affil-it]{authblk}

\usepackage{ulem}
\usepackage{graphicx}
\usepackage{amsfonts, amsmath, amsxtra, amssymb, latexsym,url}
\usepackage{makeidx}
\usepackage[utf8]{inputenc}
\usepackage[T1]{fontenc}
\usepackage[english]{babel}
\usepackage{multirow}

\usepackage{color}
\usepackage[dvipsnames]{xcolor}

\definecolor{light-blue}{rgb}{0.1,0.75,0.9} 
\definecolor{myred}{rgb}{0.8, 0.1, 0.2}

\usepackage{geometry}
\begin{document}

\title{ Solving weakly supervised regression problem using low-rank manifold regularization}
\author{Vladimir Berikov
\thanks{Email: \texttt{berikov@math.nsc.ru}}}
\affil{Sobolev Institute of Mathematics,
Novosibirsk, Russia\\
Novosibirsk State University, Novosibirsk, Russia}
\author{Alexander Litvinenko
\thanks{Email: \texttt{litvinenko@uq.rwth-aachen.de}}}
\affil{RWTH Aachen, Aachen, Germany}
\date{\vspace{-5ex}}

\maketitle





%

%
\begin{abstract}
We solve a weakly supervised regression problem. Under ``weakly'' we understand that for some training points the labels are known, for some unknown, and for others uncertain due to the presence of random noise or other reasons such as lack of resources. The solution process requires to optimize a certain objective function (the loss function), which combines manifold regularization and low-rank matrix decomposition techniques. These low-rank approximations allow us to speed up all matrix calculations and reduce storage requirements. This is especially crucial for large datasets. Ensemble clustering is used for obtaining the co-association matrix, which we consider as the similarity matrix. The utilization of these techniques allows us to increase the quality and stability of the solution. In the numerical section, we applied the suggested method to artificial and real datasets using Monte-Carlo modeling.
\end{abstract}
\textbf{Keywords:}
Weakly supervised learning, Manifold regularization, Low-rank matrix decomposition, Cluster ensemble, Co-association matrix

\tableofcontents
\section{Introduction}
Nowadays, machine learning (ML) theory and methods are rapidly developing and increasingly used in various fields of science and technology. An urgent problem remains a further improvement of ML methodology: the development of methods that allow obtaining accurate and reliable solutions in a reasonable time in conditions of noise distortions, large data size, and lack of training information. In many applications, only a small part of the data can be labeled, i.e., the values of the predicted feature are not provided for all data objects. In the case of a large amount of data and limited resources for its processing, some data objects can be inaccurately labeled. 

As a real-world example of such a problem, one can address the task of manual annotation of a large number of computed tomography (CT) digital images. In order to distinguish the brain areas affected by stroke, it is required to engage a highly qualified radiologist, and the process is rather time-consuming. It is possible that some parts of the images will stay without specifying specific regions (for example, it is simply indicated that pathological signs are present in the given CT scan) or are labeled inaccurately. In this case, the assumed region can be outlined with a frame; the closer to the center of the frame, the greater the confidence that the brain tissue is damaged. 

Weakly supervised learning is a part of ML research aimed at elaborating models and methods for the analysis of such type of information. In the formulation of a weakly supervised learning problem, it is assumed that some of the sample objects are labeled inaccurately. This inaccuracy can be understood in different ways \cite{Zhou}.

In the case of {\it coarse grained} label information, class labels are provided only for sets of objects.
For example, 
a collection of 
regulatory SNPs (Single-Nucleotide Polymorphisms) in DNA can be marked as a group linked to a pathology-associated gene. 
It is required to predict the class (its label) of each new group of objects. This task is also called {\it multi-instance learning}. This problem is considered, for example, in \cite{Xiao}, in which a modification of the SVM method is proposed for the solution. 

In another setting, it is assumed that there is an uncertainty in the indication of the exact class label arising from errors or due to the limitations of the observation method itself. Over time various solutions to the problem have been proposed. One of them is based on finding potentially erroneous labels and correcting them \cite{Muhlenbach}. A similar idea (called {\it censoring} of the sample) was developed in \cite{Borisova}. This approach usually relies on information about the nearest neighbors of points. Therefore it becomes less reliable in high-dimensional feature spaces since the points become approximately equidistant from each other. 

Another methodology is used when the labeling is performed by many independent workers (i.e., {\it crowd-sourcing} technique); among them can be both experienced and inexperienced members (and even deliberately mistaken). To solve the problem, probabilistic or ensemble methods are used \cite{Raykar,ZhouZH}.

The following approach is based on minimizing theoretical risk estimates taking into account the random labeling error. The authors of \cite{Gao} propose a method based on the fact that the empirical risk functional can be divided into two parts. The first part does not depend on the noise. Only the second part is affected by noisy labels. 

Methods based on the {\it cluster assumption} and the {\it manifold assumption} are also used \cite{Belkin,Huang}. In \cite{GaoWei}, the upper bounds of the labels' noise characteristics are obtained, and an algorithm for estimating the degree of noise is proposed using preliminary data partitioning into clusters.

In this paper, we consider a weakly supervised regression problem in the transductive learning setting. It means that the test sample is known, and the values of the predictors can be used as additional information for the target feature prediction. 

We propose a novel method using a combination of manifold regularization methodology, cluster ensemble, and low-rank matrix representation. We assume the existence of the dependence between clusters presented in the data and the predicted continuous target feature (cluster assumption). Such dependence can be found, for example, when some hidden structures are present in data, and the belonging of objects to the same structural unit affects the similarity of their target feature values.

In the rest of the paper, we give the formal statement of the problem (Sect.~\ref{sec:Descr}), describe the details of the suggested
method (Sections~\ref{sec:graph} and \ref{sec:Method}), and present the results of numerical experiments (Sect.~\ref{sec:MC}). Finally, we give some concluding remarks.

\section{Problem description and notation}
\label{sec:Descr}
Consider a dataset $ \mathbf{X} = \{x_{1},\dots,x_{n}\}$, where
$x_{i} \in \mathbb{R}^d$ is a feature vector, $d$ the dimensionality of
feature space $X=(X_1,\dots,X_d)$, and $n$ the sample size. Suppose that each data point is sampled from an unknown distribution.

Fully supervised learning assumes we are given a set $Y=\{y_1,\dots,y_{n}\}$, $y_i \in D_Y$, of target feature labels for each data point. In the regression problem, values from a continuous compact set $D_Y \subset \mathbb{R}$ are understood as target feature labels.

The objective is to find a decision function $y=f(x)$, which should forecast the target feature for new examples from the same distribution. The decision function should optimize a quality metric, e.g., minimize an estimate of the expected loss.

In an unsupervised learning problem, the target feature is not specified. 
It is necessary to find a meaningful representation of data, i.e., find a
partition $P = \{C_1,\dots, C_K \}$ of $\mathbf{X}$ on a relatively
small number $K$ of homogeneous clusters describing the structure of
data. The homogeneity criterion depends on the similarity of observations within clusters and the distances between them. Quite often, the optimal number of clusters is unknown and should be determined using a cluster validity index.

The obtained cluster partition can be uncertain due to a lack of knowledge about data structure, vagueness in setting optional parameters of the learning algorithm, or dependence on random initializations. In this case, ensemble clustering is a way of obtaining a robust
clustering solution. This methodology aims at finding consensus partition
from different partition variants \cite{Boongoen}. A properly organized
ensemble (even composed of  ``weak'' algorithms) often significantly
improves the clustering quality.

In the problem of semi-supervised transductive learning, the target
feature labels are known only for a part of the data set  $\mathbf{X}_1\subset \mathbf{X}$ (of comparatively small size as usual). We assume that $\mathbf{X}_1=\{x_{1},\dots,x_{n_1}\}$, and the unlabeled part is
$\mathbf{X}_0=\{x_{n_1+1},\dots,x_{n}\}$. The set of labels for
points from $\mathbf{X}_1$ is denoted by
$\mathbf{Y}_1=\{y_1,\dots,y_{n_1}\}$. It is required to predict labels $\mathbf{Y}_0=(y_{n_1+1},\dots,y_n)$ in the best way for the unlabeled sub-sample $\mathbf{X}_0$. 

This task is essential because in many applied problems only a small part of available data can be labeled due to the considerable cost of target feature registration. 

We consider a weakly supervised learning context, i.e., we suppose that for some data points, the labels are known, for some unknown, and for others uncertain due to reasons such as lack of resources for their careful labeling or presence of random distortions arising in the label identification process.  

To model the uncertainty in the label identification, we suppose that for each $i$th data point, $i=1,\dots,n_1$, the value $y_i$ of the target feature is a realization of a random variable $Y_i$ with cumulative distribution function  (cdf) $F_i(y)$ defined on $D_Y$. We suppose that $F_i(y)$ belongs to a given distribution family.

In this paper, the regression problem is considered, i.e., the predicted feature is continuous. 
Further we assume the following normal distribution model for the uncertain target variable:  
\begin{equation}
\label{Fi}
Y_i \sim \mathcal{N}(a_i,\sigma_i),
\end{equation}
where $a_i,\sigma_i$ are the mean and the standard deviation respectively. The larger $\sigma_i$, the more uncertain is the labelling. It is presumed that parameters $a_i=y_i$ and $\sigma_i=s_i$ are known for each (weakly) labeled observation, $i=1,\dots,n_1$.  For strictly determined observation $y_i$, we nevertheless postulate a normal uncertainty model with $a_i=y_i$ and small standard deviation $\sigma_i \approx 0$.


We aim at finding a weak labeling of points from $\mathbf{X}_0$, i.e., determining $F_i(y)$ for ${i=n_1+1,\dots,n}$ following an objective criterion. 

\section{Manifold regularization}
\label{sec:graph}

Semi-supervised learning and weakly supervised learning assume two basic assumptions: cluster assumption and the assumption that the data with similar labels belong to a low-dimensional manifold. 

According to the cluster assumption, one believes that objects from the same cluster often have the same labels or labels close to each other. 

The manifold assumption is based on the hypothesis that there is a smooth manifold (for example, a two-dimensional surface in multidimensional space) to which points with similar labels belong. 
Manifold regularization  \cite{Belkin,van2020survey} is based on this assumption. In addition to the learning error, the regularizing component is minimized during the model fitting stage. The component characterizes the smoothness of the decision function change. In dense regions, the decision function must change slowly, so its gradient must be small. In other words, data points from $\mathbf{X}$ lie on a low-dimensional non-linear manifold $M$, and the decision function is smooth on this manifold, i.e., points close to each other possess similar labels.


In semi-supervised learning, the regularization functional to be minimized can be written as following:
$$
J(f)= \frac{1}{n_1}\sum\limits_{x_{i} \in X_{1}} V (y_{i}
,f_{i}) + \gamma \, \left\| f \right\|^2_M,
$$
where $f=(f_1,\dots,f_n)$ is a vector of predicted labels, $V$ a loss function, $ \gamma >0$ a regularization parameter, and  $\left\| f \right\|_M^2$ characterizes the smoothness of the function. In dense regions, the decision function should change slowly, i.e., its gradient  $\nabla _{M}f(x)$ should be small. Thus, the manifold regularizer can be chosen in this way:
$$\left\| f \right\|^2_M =\int _{x\in M}\left\|\nabla _{M}f(x)\right\|^{2}\,d{\mathcal {P}}_{X}(x).$$

Graph Laplacian (GL) \cite{Belkin,ZhouD} is a convenient tool for $\left\| f \right\|^2_M$ estimation.
Let $ G = (V, E) $ be a weighted non-oriented complete graph, in
which the set of vertices $ V $ corresponds to points from $
\mathbf{X} $, and the set of edges $ E $ corresponds to pairs $
(x_{i}, x_{j}), \, \, i, j = 1, \dots, n, \, \, \, i \ne j $. Each
edge $ (x_{i}, x_{j}) $ is associated with a non-negative weight $
W_{ij} $ (the degree of similarity between the points). 

The degree of similarity can be calculated by using an appropriate function, for
example from the Mat\'{e}rn family \cite{Matern1986a}. The
Mat\'{e}rn function depends only on the distance $h:=\Vert
x_i-x_j\Vert $ and is defined as
$$W(h)=\frac{\sigma^2}{2^{\nu-1}\Gamma(\nu)}\left(\frac{h}{\ell}\right)^\nu K_\nu\left(\frac{h}{\ell}\right)$$
with three parameters $\ell$, $\nu$, and $\sigma^2$. For instance,
$\nu=1/2$ gives the well-known exponential kernel
$W(h)=\sigma^2\exp(-h/\ell)$, and $\nu \rightarrow \infty$ gives the Gaussian
kernel $W(h)=\sigma^2\exp(-h^2/2\ell^2)$.
In this paper we use 
the Gaussian covariance function, also called the radial basis function (RBF kernel), with $\sigma=1$:
\begin{equation}
\label{eq:RBF}
W_{ij}=\exp\left(-\frac{\Vert x_i-x_j\Vert^2}{2 \ell^2}\right).
\end{equation}

By $L:=D-W$ we denote the standard GL, where $D$ is a diagonal
matrix with elements $D_{ii} = \sum\limits_{j} W_{ij}$.  There are also normalized GL: $L_{norm} = D^{- 1/2} W \, D^{- 1/2}$ and the random walk GL: $L_{rw}=D^{-1}W$.

One can show that in semi-supervised regression, the regularization term can be expressed as following:
$$
 \left\| f \right\|^2_M \approx  \frac{1}{n^2} f^\top L f =  \frac{1}{n^2} \sum\limits_{x_i,x_j \in
\mathbf{X}} W_{ij} (f_i-f_j)^2.
$$

\section{Proposed method}
\label{sec:Method}

Consider a modification of the manifold regularization scheme for a considered weakly supervised transductive learning problem. 

Let $F=\{F_1,\dots,F_n\}$ denote the set of cdfs for data points; each cdf $F_i$ is represented by a pair of parameters $(a_i, \sigma_i)$.

\subsection{Objective functional}

Consider the following optimization problem: 
\vskip12pt

\begin{center}
    find $F^{*}= \arg\min\limits_{F} J(F)$, where 
\end{center} 
\begin{equation}
\label{QF}
J(F)=
\sum\limits_{x_i \in X_1}
\mathcal{D}(Y_i,F_i) +
\gamma \sum\limits_{x_i,x_j \in \mathbf{X}}
\mathcal{D}(F_i,F_j) W_{ij}.
\end{equation}
Here $\mathcal{D}$ is a statistical distance between two distributions (such as the Wasserstein distance, Kullback-Leibler divergence, or other metrics). 
The first sum in the right side of (\ref{QF}) is aimed to reduce the dissimilarity on labeled data; the second component plays the role of a smoothing function: its minimization means that if two points $x_{i}, x_{j} $ (either labeled or unlabeled) are similar, their labeling distribution should not be very different. 

In this work, we use the Wasserstein distance $w_p$ \cite{Bogachev} (also known as the Kantorovich-Rubinstein distance or transportation metric) between distributions $P$ and $Q$ over a set $D_Y$ as a measure of their dissimilarity:
$$
w_{p}(P,Q ):=\left(\inf_{\gamma \in \Gamma (P ,Q )}\int _{D_Y\times D_Y} \rho (y_1,y_2)^{p}\,\mathrm {d} \gamma (y_1,y_2)\right)^{1/p},
$$
where $ \Gamma (P ,Q )$ is a set of all probability distributions on $D_Y\times D_Y$ with marginal distributions $P $ and $Q$, $\rho$ a distance metric, and $p \ge 1$.

It is known that for normal distributions $P_i=\mathcal{N}(a_i,\sigma_i)$, $Q_j=\mathcal{N}(a_j,\sigma_j)$ and the Euclidean metric, the $w_2$ distance is equal to \cite{Delon}
$$w_{\text{2}}(P_i, Q_j)=(a_i-a_j)^2+(\sigma_i-\sigma_j)^2.$$
We use the $w_2$ distance in (\ref{QF}) for weakly supervised regression and slightly modify the objective functional in (\ref{QF}) adding an $L_2$ regularizer:
$$J(a,\sigma)=\sum\limits_{i=1}^{n_1} \left( (y_i-a_i)^2 + (s_i-\sigma_i)^2 \right)+$$
\begin{equation}
\label{OptREg}
+ \gamma \sum\limits_{i,j=1}^{n} \left(  (a_i-a_j)^2+(\sigma_i-\sigma_j)^2    \right) W_{ij} + \beta (\left\| a \right\|^2 + \left\| \sigma \right\|^2),
\end{equation}
where $\beta > 0$ is a regularization parameter, $a=(a_1,\dots,a_n)^\top$, $\sigma=(\sigma_1,\dots,\sigma_n)^\top$. 

\subsection{Optimal solution}
To find the optimal solution, we differentiate (\ref{OptREg}) and get:
\begin{equation} \label{Q2}
\frac {\partial J} {\partial a_i} =
2(a_i-y_i)+4\gamma \sum\limits_{j=1}^{n} (a_i-a_j) W_{ij} +2 \beta a_i  = 0, \, \, \, \, i = 1, \dots, n_{1},
\end{equation}
\begin{equation}\label {Q3}
\frac {\partial J} {\partial a_i} =
4\gamma \sum\limits_{j=1}^{n} (a_i-a_j) W_{ij} + 2 \beta a_i   = 0, \, \, \, \,
\, i = n_{1} +1, \dots, n.
\end{equation}
Denote $Y_{1,0}=(y_1,\dots,y_{n_1},
\underbrace{0,\dots,0}_{n-n_1})^\top$ and let $B$ be a diagonal matrix
with elements 
$$ B_{ii} =\left\{^{\beta+1, \; i=1,\dots,n_1}_{\beta,
\; i=n_1+1,\dots,n.} \right.$$
Combining (\ref{Q2}), (\ref{Q3}) and using vector-matrix notation, we finally get:
$$ (B+ 2 \gamma L ) a = Y_{1,0},$$ thus the optimal solution is
\begin{equation}
\label{a*}a^{*} = ( B +2 \gamma L)^{-1} Y_{1,0} .\end{equation}

Similarly, one can obtain the optimal value of $\sigma$:
\begin{equation} 
\label{sig*} \sigma^{*} = ( B + 2 \gamma L)^{-1} S_{1,0}, 
\end{equation}
where $S_{1,0}=(s_1,\dots,s_{n_1},
\underbrace{0,\dots,0}_{n-n_1})^\top$.

\subsection{Low-rank similarity matrix representation} 
For large-scale problems, the dimensionality of matrices to be inverted in (\ref{a*}), (\ref{sig*}) is very large and the inversion is costly. In many applications, a low-rank matrix decomposition is a useful tool for obtaining computationally efficient solutions \cite{GH03}. Nystr{\"o}m method (see, e.g., \cite{Drineas}) or hierarchical low-rank matrix approximations \cite{Part1,HackHMEng,khoromskij2009application,LitvGentonSunKeyes17,litvHLIBPro17,litv17Tensor}
can be used for obtaining such a decomposition.

Let the similarity matrix be presented in the low-rank form
\begin{equation}
\label{lr}
W = A A^\top,
\end{equation}
where matrix $A \in \mathbb{R}^{n\times m}$, $m \ll n$. Further, we have
\begin{equation} \label{B}
B + 2 \gamma L = B + 2 \gamma D - 2 \gamma A A^\top = G - 2 \gamma A A^\top,
\end{equation}
where $ G = B + 2 \gamma D$. 

The following Woodbury matrix identity is well-known in linear algebra:
\begin{equation} \label{WoodB}
(S + UV)^{-1} = S^{-1} - S^{-1} U(I + V S^{-1} U)^{-1} VS^{-1},
\end{equation}
where $S \in \mathbb{R}^{n\times n}$ is an invertible matrix, $U \in
\mathbb{R}^{n\times m}$ and $V \in \mathbb{R}^{m\times n}$. 

Let $S=G$, $U=-2 \gamma A$ and $V=A^\top$. One can see that
\begin{equation} \label{SG}
G^{-1}=\text{diag}\left(1/(B_{11}+2 \gamma D_{11}), \dots,1/(B_{nn} + 2 \gamma D_{nn})\right).
\end{equation}
From (\ref{a*}), (\ref{B}), (\ref{WoodB}) and  (\ref{SG}) we obtain:
\begin{equation} \label{alr}
a^{*}=(G^{-1}+2 \gamma G^{-1}A (I-2 \gamma A^\top
G^{-1}A)^{-1}A^\top G^{-1})\;Y_{1,0}.
\end{equation}

Similarly, from (\ref{sig*}), (\ref{B}), (\ref{WoodB}) and  (\ref{SG}) we have:
\begin{equation} \label{slr}
\sigma^{*}=(G^{-1}+2 \gamma G^{-1}A (I-2 \gamma A^\top
G^{-1}A)^{-1}A^\top G^{-1})\;S_{1,0}.
\end{equation}

Note that in (\ref{alr}) and (\ref{slr}) one needs to invert a matrix of significantly smaller dimensionality
$m \times m$  instead of $n \times n$ matrix in (\ref{a*}) and (\ref{sig*}).
The computational complexity of (\ref{alr}) and (\ref{slr}) can be
estimated as $O(nm+m^3)$.

\subsection{Co-association matrix of cluster ensemble}
\label{sec:Co-ass}
We use a co-association matrix of cluster
ensemble as a similarity matrix in (\ref{OptREg}) \cite{BerikovS}. The co-association matrix
is calculated in the process of cluster ensemble creation.

Let us consider a set of partition variants $\{P_{l}\}_{l=1}^r$,
where $P_l = \{C_{l,1} ,\dots,C_{l,K_l} \}$, $C_{l,k} \subset
\mathbf{X}$, $C_{l,k}\bigcap C_{l,k'}=\varnothing$ and $K_l$ is the number
of clusters in $l$th partition. For each partition $P_l$ we determine matrix
$H_l=(h_l(i,j))_{i,j=1}^n$ with elements indicating whether a pair
$x_i$, $x_j$ belong to the same cluster in $l$th variant or not.
We have 
$$h_l(i,j)=\mathbb{I}[c_l(x_{i} ) = c_l(x_{j} )],$$
where $\mathbb{I}(\cdot )$ is the indicator function with $\mathbb{I}[true]=1$,
$\mathbb{I}[false]=0$, and $c_l(x)$ is the cluster label assigned to $x$.
The weighted averaged co-association matrix is
\begin{equation}
\label{eq:defH}
H =  \sum\limits_{l=1}^r \omega_l H_l,
\end{equation}
where $\omega_1,\dots,\omega_r$ are weights of ensemble
elements, $\omega_l \ge 0$, $\sum \omega_l=1$. The weights are used to assess the
importance of base clustering variants \cite{Berikov2017}. They
depend on the evaluation function $\Gamma$ (e.g., cluster validity
index) \cite{Berikov2017}: $\omega_l=\gamma_l /
\sum\limits_{l'} \gamma_{l'}$, where $\gamma_l=\Gamma(l)$ is an
estimate of the clustering quality for the $l$th partition.

The matrix $H$ can be considered as a pairwise similarity matrix which determines the similarity between points in a new feature
space obtained with an implicit transformation of data. 

It is easy to see that $H$ admits a low-rank decomposition in the form:
\begin{equation}
\label{H} H = R R^\top,
\end{equation}
where $R=[R_1 R_2 \dots R_r]$, $R$ is a block matrix, $R_l=\sqrt{\omega_l}\, Z_l$, $Z_l$ is ($n
\times K_l$) cluster assignment matrix for $l$th partition:
$Z_l(i,k)=\mathbb{I}[c(x_i)=k]$, $i=1,\dots,n$, $k=1,\dots,K_l$.


As a rule, $m=\sum_l K_l \ll n$, thus (\ref{H})
gives us an opportunity of saving memory by storing $(n \times m)$
sparse matrix instead of full $(n \times n)$ co-association matrix.
%

The Graph Laplacian matrix for $H$ can be written in the form: $$L=D'-H,$$ where
$D'=\text{diag}(D'_{11},$ $\dots, D'_{nn})$, $D'_{ii} =
\sum\limits_{j} H(i,j)$. One can see that

\begin{equation}
\label{D'} D'_{ii} = \sum\limits_{j=1}^{n} \sum\limits_{l=1}^{r} \omega_l
\sum\limits_{k=1}^{K_l} Z_l(i,k) Z_l(j,k) 
=\sum\limits_{l=1}^{r} \omega_l N_{l}(i),
\end{equation}
\noindent where $N_{l}(i)$ is the size of the cluster which includes
point $x_i$ in $l$th partition variant.

Using $H$ in the low-rank representation (\ref{H}) instead of the similarity matrix $W$ in (\ref{lr}), and the matrix $D'$ defined in (\ref{D'}), we obtain cluster ensemble based predictions in the form given by (\ref{alr}), (\ref{slr}).

\subsection{WSR-LRCM algorithm}
The basic steps of the suggested weakly supervised regression algorithm based on the low-rank representation of the co-association matrix (WSR-LRCM)
are as follows.

\bigskip
\noindent\textbf{Input}:

\noindent ${ {\mathbf X}}$: dataset including both labeled,  inaccurately labeled and unlabeled samples;

\noindent $a_i$, $\sigma_i$, $i=1,\dots,n_1$: uncertain input  parameters for labeled and inaccurately labeled points;

\noindent $r$, $\mathbf{\Omega}$ : number of runs and set of parameters for the $k$-means clustering (number of clusters, maximum number of iterations, parameters of the initialization process).

\noindent\textbf{Output}:

\noindent $a^{*}$, $\sigma^{*}$: predicted estimates of uncertain parameters for objects from  sample $\mathbf X$ (including predictions for the unlabeled sample).

\noindent\textbf{Steps:}

\noindent 1. Generate $r$ variants of clustering partition for  parameters randomly chosen from
$\mathbf{\Omega}$; calculate weights $\omega_1,\dots,\omega_r$ .

\noindent 2. Find a low-rank representation of the graph Laplacian using (\ref{H}) and $D'$ in (\ref{D'});

\noindent 3. Calculate predicted estimates of  uncertainty parameters using (\ref{alr}) and (\ref{slr}) .

\noindent\textbf{end.}

%
\section{Monte-Carlo experiments}
\label{sec:MC}
This section presents the results of numerical experiments with the proposed WSR-LRCM algorithm. The regression quality and running time are experimentally evaluated
on two synthetic examples (1a and 1b) and one real example.

\subsection{Settings in Examples 1a and 1b with artificial data}
In both examples we consider datasets generated from a mixture
of two multidimensional normal distributions
$\mathcal{N}(m_1,\sigma_X I)$, $\mathcal{N}(m_2,\sigma_X I)$ with
equal weights; $m_1$, $m_2$ $\in \mathbb{R}^d$, $d=8$, $\sigma_X$
is a parameter.

Let the ground truth target feature is equal to $Y=1+\varepsilon$ for points generated from
the first component, otherwise $Y=2+\varepsilon$, where
$\varepsilon$ is a normally distributed random value with zero mean and variance
$\sigma^2_\varepsilon$. 

During Monte Carlo simulations, we generate samples of the given size $n$
according to the specified distribution mixture. Two-thirds of the sample points are included into the training part $\mathbf{X}_{train}$, and the remaining points compose the test sample $\mathbf{X}_{test}$. In the training sample, 10\%
of the points selected at random from each component comprise a fully
labeled sample; 20\% of the sample consists of inaccurately labeled objects; the remaining part contains the unlabeled data. Such partitioning mimics a typical situation in the weakly supervised learning: a small number of accurately labeled instances, medium sized uncertain labelings and a lot of unlabeled examples. To model the inaccurate labeling, we use the parameters defined in (\ref{Fi}):
\begin{equation}
\label{eq:delta}
\sigma_i=\delta \cdot \sigma_Y,    
\end{equation}
where $\sigma_Y$ is a standard deviation of $Y$ over labeled data, and $\delta > 0$ is a parameter.

The quality of prediction is estimated on the test sample as the mean Wasserstein distance between the predicted, according to (\ref{alr}) and (\ref{slr}), and ground truth values:
\begin{equation}
\label{eq:MWD}
\text{MWD}=\frac{1}{n_{test}}\sum\limits_{x_i \in \mathbf{X}_{test}} \left( (a_i^{true}-a^{*}_i)^2+ \sigma^{*2}_i \right),
\end{equation}
where $n_{test}$ is test sample size, $a_i^{true}=y_i^{true}$ is the true value of the target feature. Note that the standard Mean squared error (MSE) quality metric can be considered as a special case of MWD for accurate labeling.

The ensemble size is $r=10$.
The weights of ensemble elements are the same: $\omega_l\equiv 1/r$. The
regularization parameters $\beta, \, \gamma$, used in (\ref{Q2}), have been estimated using the
grid search and cross-validation techniques. In our experiments, the
best results were obtained for $\beta=0.001$ and $\gamma=0.001$.

Further, we compare the suggested WSR-LRCM method with its simplified version,
the semi-supervised regression algorithm SSR-RBF considered in our previous work \cite{BL19},
which uses the standard similarity matrix evaluated with the RBF
kernel as in (\ref{eq:RBF}).  The output predictions were
calculated according to (\ref{a*}) and (\ref{sig*}). Due to the fact that in semi-supervised learning only labeled and unlabeled instances can be used, the SSR-RBF algorithm considers inaccurately labeled objects as unlabeled. The same RBF kernel is applied in both algorithms. The parameters of the data generation procedure remain unchanged, $n=1000$, $\sigma_\varepsilon = 0.1$. The MWD metric is utilized for quality evaluation in both cases.

To increase the statistical reliability of the results, we average the obtained estimates over 40 Monte
Carlo repetitions (except cases with $n \ge 10^5$). All algorithms were run on a dual-core Intel Core i5 processor with the
clock frequency of 2.4 GHz and 8 GB RAM.
The mean values are
$m_1=(0,\dots,0)^\top$ and $m_2=(10,\dots,10)^\top$.

\subsection{Additional settings and results of Example 1a}
The ensemble variants are generated by random initialization of centroids
(number of clusters equals two). Different values of parameter $\ell$ were considered, and a quasi-optimal $\ell= 6.6$ was determined.

Table~\ref{T1} shows the averaged values of the MWD metric and computing times for different sample sizes and values of parameter $\sigma_{\varepsilon}$. In the data generation procedure $\sigma_X=2$ and $\delta=0.1$ were used.

\begin{table}[h!] \centering
\caption{
Comparison of averaged MWD estimates and running times for WSR-LRCM and WSR-RBF algorithms.}
\label{T1}
\begin{center}
\renewcommand{\arraystretch}{1.3} 
\begin{tabular}{|c|c|c|c|c|c|}  \hline
  \multirow{2}{36pt}{\hskip12pt $n$} &\multirow{2}{24pt}{$\;\;\sigma_\varepsilon$}
  & \multicolumn{2}{|c|}{WSR-LRCM} & \multicolumn{2}{c|}{WSR-RBF}
  \\
  \cline{3-6} & & \; MWD \; & time (sec)\;
   & \;MWD\; & \;  time  (sec)\;  \\
\hline \multirow{3}{36pt}{\;\;1000}
 & 0.01 & 0.0015 & 0.02  & 0.0027 &  0.045  \\
 & 0.1 & 0.012 & 0.02 & 0.013 &  0.046  \\
 & 0.25 & 0.065 & 0.02  & 0.066 &  0.047  \\
 \hline \multirow{3}{36pt}{\;\;5000}
 & 0.01 & 0.0013 & 0.03  & 0.0014 &  1.79  \\
 & 0.1 & 0.011 & 0.03 & 0.011 &  1.75 \\
& 0.25 & 0.064 & 0.03  & 0.064 &  1.8 \\
\hline \multirow{3}{36pt}{\;\;10000}
 & 0.01 & 0.0013 & 0.05 & 0.0013 & 9.9  \\
 & 0.1 & 0.011 & 0.05 & 0.011 &  9.4  \\
& 0.25 & 0.063 & 0.05 &  0.063 &  9.9  \\
 \hline \multirow{1}{36pt}{\;\;\;$10^5$}
 & 0.01 & 0.0013 & 0.67 &  - &  -  \\
  \hline \multirow{1}{36pt}{\;\;\;$10^6$}
 & 0.01 & 0.0013 & 7.55 & - &  -  \\
  \hline \multirow{1}{36pt}{\;\;\;$10^7$}
 & 0.01 & 0.0013 & 100 & - &  -  \\
 \hline
\end{tabular}
\end{center}
\end{table}
One can see that WSR-LRCM produces similar results as WSR-RBF with respect to the MWD metric. At the same
time,  WSR-LRCM runs much faster. For large sample sizes ($n \ge 10^5$) the WSR-RBF method fails due to unacceptable memory demand.

In Table~\ref{T2}, we investigate performance of both algorithms for different values of the parameter $\delta$. Again, the WSR-LRCM algorithm shows smaller MWD errors.

\begin{table}[h!] \centering
\caption{Comparison of averaged MWD estimates for different values of $\delta$, defined in (\ref{eq:delta}).} \label{T2}
\begin{center}
\renewcommand{\arraystretch}{1.3} 
\begin{tabular}{|c|c|c|c|}  \hline
  $\delta$ & 0.1 & 0.2  & 0.3   \\ \hline
  \; WSR-LRCM \;& \;0.010 \; & \;0.015 \;& \; 0.020 \;  \\ \hline
  \;SSR-RBF \; & \; 0.022 \; & \; 0.022 \; & \; 0.022\;    \\ 
 \hline
\end{tabular}
\end{center}
\end{table}

\subsection{Additional settings and results of Example 1b}
This example is more complicated. Below we describe what is different. To investigate the robustness of the algorithm, we added
noise to the data by appending two independent features of a uniform distribution $\mathcal{U}(0,1)$.

To increase the diversity of base clusterings, we set the number of clusters in each run as $K=2,\dots, K_{max}$, where $K_{max}=2+r$.
A quasi-optimal $\ell= 1.85$ was determined.

Table~\ref{T3} shows the averaged values of the MWD metric and computing times for different sample sizes and values of parameter $\sigma_{\varepsilon}$. In the data generation procedure $\sigma_X=3$ (larger overlap of clusters in comparison to Example 1a) and $\delta=0.1$ were used.

\begin{table}[h!] \centering
\caption{
Comparison of averaged MWD estimates and running times for WSR-LRCM and WSR-RBF algorithms.}
\label{T3}
\begin{center}
\renewcommand{\arraystretch}{1.3} 
\begin{tabular}{|c|c|c|c|c|c|}  \hline
  \multirow{2}{36pt}{\hskip12pt $n$} &\multirow{2}{24pt}{$\;\;\sigma_\varepsilon$}
  & \multicolumn{2}{|c|}{WSR-LRCM} & \multicolumn{2}{c|}{WSR-RBF}
  \\
  \cline{3-6} & & \; MWD \; & time (sec)\;
   & \;MWD\; & \;  time  (sec)\;  \\
\hline \multirow{3}{36pt}{\;\;1000}
 & 0.01 & 0.002 & 0.04  & 0.007 &  0.04  \\
 & 0.1 & 0.012 & 0.04 & 0.017 &  0.04  \\
 & 0.25 & 0.065 & 0.04  & 0.070 &  0.04  \\
 \hline \multirow{3}{36pt}{\;\;5000}
 & 0.01 & 0.001 & 0.14 & 0.004 &  1.71  \\
 & 0.1 & 0.011 & 0.14 & 0.014 &  1.72 \\
& 0.25 & 0.064 & 0.15  & 0.067 &  1.75 \\
\hline \multirow{3}{36pt}{\;\;10000}
 & 0.01 & 0.001 & 0.33 & 0.002 & 9.40  \\
 & 0.1 & 0.011 & 0.33 & 0.012 &  9.35  \\
& 0.25 & 0.064 & 0.33 &  0.065 &  9.36  \\
 \hline \multirow{1}{36pt}{\;\;\;$10^5$}
 & 0.01 & 0.001 & 6.72 &  - &  -  \\
  \hline \multirow{1}{36pt}{\;\;\;$10^6$}
 & 0.01 & 0.001 & 89.12 & - &  -  \\
 \hline
\end{tabular}
\end{center}
\end{table}
Table~\ref{T3} demonstrates that WSR-LRCM produces nearly the same or better results than WSR-RBF with respect to the WMD metric. At the same
time,  WSR-LRCM runs much faster. For large sample sizes ($n \ge 10^5$) the WSR-RBF method fails due to unacceptable memory demand.
%
In Table~\ref{T4}, we investigate performance of both algorithms for different values of the parameter $\delta$. Again, the WSR-LRCM algorithm shows smaller MWD errors.

\begin{table}[h!] \centering
\caption{Comparison of averaged MWD estimates for different values of $\delta$, defined in (\ref{eq:delta}).}
\label{T4}
\begin{center}
\renewcommand{\arraystretch}{1.3} 
\begin{tabular}{|c|c|c|c|}  \hline
  $\delta$ & 0.1 & 0.25  & 0.5   \\ \hline
  \; WSR-LRCM \;& \;0.012 \; & \;0.017 \;& \; 0.038 \;  \\ \hline
  \;SSR-RBF \; & \; 0.051 \; & \; 0.051 \; & \; 0.051\;    \\ 
 \hline
\end{tabular}
\end{center}
\end{table}
From Tables~\ref{T2} and \ref{T4} one may conclude that additional information on uncertain labelings improves the quality of forecasting in WSR-LRCM. The parameter $\delta$ accounts for the degree of uncertainty: the larger its value is, the more similar become the results of weakly supervised and semi-supervised regression. 

\subsection{Example 2 with real data }

We analyse the Gas Turbine CO and NOx Emission Data Set \cite{http,kaya2019predicting}. 
This dataset includes measurements of 11 features describing working characteristics (temperature, pressure, humidity, etc.) of a gas turbine located in Turkey. The monitoring was carried out during 2011-2015. Carbon monoxide (CO) and
Nitrogen oxides (NOx) are the predicted outputs. The forecasting of these  harmful pollutants is necessary for controlling and reducing the emissions from power
plants.

We make predictions for CO over the year 2015 (in total, 7384 observations) and use the following experiment's settings. The dataset is randomly partitioned on learning and test samples in the proportion 2:1. The volume of the accurately labeled sample is 1\% of overall data; 10\% of data are considered as inaccurately labeled instances; the remaining data are regarded as unlabeled samples. As in the previous example, we use the $k$-means clustering as the base ensemble algorithm (the number of clusters varies from 100 to $100+r$). All other settings are the same. 

As a result of forecasting, the averaged MWD for WSR-LRCM
takes the value $1.85$ and for SSR-RBF the value $5.18$.  

In order to compare WSR-LRCM with fully supervised algorithms, we calculate the standard Mean Absolute Error (MAE) using estimates of $a^*$ defined in (\ref{alr}) as the predicted feature outputs:
$$\text{MAE}=\frac{1}{n_{test}}\sum\limits_{x_i \in \mathbf{X}_{test}} |y_i^{true}-a^{*}_i |.$$

The Random Forest (RF) and Linear Regression (LR) methods are evaluated taking accurately labeled examples as the learning sample. 
Table~\ref{T5} summarize comparison of three methods: WSR-LRCM, RF and LR.

\begin{table}[h!] \centering
\caption{Comparison of WSR-LRCM, RF and LR methods}
\begin{center}
\renewcommand{\arraystretch}{1.3} 
\label{T5}
\begin{tabular}{|c|c|c|}  \hline
   & averaged MAE & averaged computing time (sec)    \\ \hline
  WSR-LRCM & 0.634  & 1.99    \\ \hline
  RF(with 300 trees) & 0.774  &  0.35  \\ \hline 
  LR & 0.873  &  0.38  \\ \hline
\end{tabular}
\end{center}
\end{table}

A growth in the computing time for WSR-LRCM in this experiment can be explained by a large number of clusters ($> 100$), which is one of the parameters of the $k$-means clustering method. 

From these experiments, one may conclude that the proposed WSR-LRCM method gives more accurate predictions than other compared methods in case when proportion of the labeled sample is small.

\section*{Conclusion}
In this work, we have introduced a weakly supervised regression method using the manifold regularization technique. We have considered the case where the learning sample includes labeled, unlabeled and inaccurately labeled instances. To model the uncertain labeling, we have used the normal distribution with different parameters. The measure of similarity between uncertain labelings was formulated in terms of the Wasserstein distance between probability distributions.

Two variants of the algorithm were proposed:  WSR-RBF, which is based on the standard RBF kernel, and  WSR-LRCM, which uses a
low-rank representation of the
co-association matrix of the cluster ensemble.
The reason for this modification is that the low-rank decomposition reduces the memory requirement and computing time. 
The ensemble clustering allows a better discovering of more complex data structures under noise distortions.
The co-association matrix depends on the decisions of clustering
algorithms and is less noise-dependent than standard
similarity matrices. 

The efficiency of the suggested methods was studied
experimentally. 

In all experiments, the proposed WSR-LRCM method demonstrated a more accurate prediction than other compared methods.

In all tests, the WSR-LRCM method has shown 
smaller running time in comparison with 
WSR-RBF. It has been shown that taking into consideration additional information on uncertain labelings improves the regression quality. 

In the future, we plan to improve our method by using deep learning methodology (in particular, deep autoencoder) at the stage of ensemble clustering. It would be interesting to investigate different variants of hierarchical low-rank decomposition techniques. Applications of this method in various fields are also
planned, especially for the analysis of computed tomography images and studying the relationships between single nucleotide polymorphisms in DNA sequences.

\section*{Acknowledgements}

The study was carried out within the framework of the state contract of the Sobolev Institute of Mathematics (project no 0314-2019-0015). The work was partly supported by RFBR grants 19-29-01175 and 18-29-09041. A. Litvinenko was supported by funding from the Alexander von Humboldt Foundation.

\bibliographystyle{splncs04}
\bibliography{motor21.bib}

\begin{thebibliography}{10}
\providecommand{\url}[1]{\texttt{#1}}
\providecommand{\urlprefix}{URL }
\providecommand{\doi}[1]{https://doi.org/#1}

\bibitem{http}
{UC Irvine Machine Learning Repository: Gas Turbine CO and NOx Emission Data
  Set}.
  \url{https://archive.ics.uci.edu/ml/datasets/Gas+Turbine+CO+and+NOx+Emission+Data+Set}
  (06 Apr 2021)

\bibitem{Belkin}
Belkin, M., Niyogi, P., Sindhwani, V.: Manifold regularization: A geometric
  framework for learning from labeled and unlabeled examples. Journal of
  Machine Learning Research  \textbf{7}(85),  2399--2434 (2006),
  \url{http://jmlr.org/papers/v7/belkin06a.html}

\bibitem{Berikov2017}
Berikov, V.B.: Construction of an optimal collective decision in cluster
  analysis on the basis of an averaged co-association matrix and cluster
  validity indices. Pattern Recognition and Image Analysis  \textbf{27}(2),
  153--165 (2017). \doi{10.1134/S1054661816040040}

\bibitem{BL19}
Berikov, V., Litvinenko, A.: Semi-supervised regression using cluster ensemble
  and low-rank co-association matrix decomposition under uncertainties.
  Proceedings of 3rd Int. Conf. on Uncertainty Quantification in CSE pp.
  229--242 (2020). \doi{10.7712/120219.6338.18377},
  \url{https://files.eccomasproceedia.org/papers/e-books/uncecomp_2019.pdf}

\bibitem{BerikovS}
Berikov~V., Karaev~N., T.A.: Semi-supervised classification with cluster
  ensemble. In: Engineering, Computer and Information Sciences (SIBIRCON), 2017
  International Multi-Conference. pp. 245--250. IEEE, Novosibirsk (2017)

\bibitem{Bogachev}
Bogachev, V.I.;~Kolesnikov, A.: {The Monge-Kantorovich problem: achievements,
  connections, and perspectives}. Russian Math. Surveys.  \textbf{67},
  785--890 (2012)

\bibitem{Boongoen}
Boongoen, T., Iam-On, N.: Cluster ensembles: A survey of approaches with recent
  extensions and applications. Computer Science Review  \textbf{28},  1--25
  (2018). \doi{https://doi.org/10.1016/j.cosrev.2018.01.003},
  \url{https://www.sciencedirect.com/science/article/pii/S1574013717300692}

\bibitem{Borisova}
Borisova, I.A., Zagoruiko, N.: Algorithm {FR}i{S}-{TDR} for generalized
  classification of the labeled, semi-labeled and unlabeled datasets. In:
  Aleskerov, F., Goldengorin, B., Pardalos, P. (eds.) Clusters, Orders, and
  Trees: Methods and Applications. Springer Optimization and Its Applications,
  vol.~92, pp. 151--165. Springer, New York, NY (2014).
  \doi{10.1007/978-1-4939-0742-7$\_$9}

\bibitem{Delon}
Delon, J., Desolneux, A.: A {W}asserstein-type distance in the space of
  {G}aussian mixture models. SIAM Journal on Imaging Sciences  \textbf{13}(2),
  936--970 (2020). \doi{10.1137/19M1301047}

\bibitem{Drineas}
Drineas, P., Mahoney, M.W., Cristianini, N.: On the nyström method for
  approximating a gram matrix for improved kernel-based learning. Journal of
  Machine Learning Research  \textbf{6},  2153--2175 (2005)

\bibitem{Gao}
Gao, W., Wang, L., li, Y.F., Zhou, Z.H.: Risk minimization in the presence of
  label noise. Proceedings of the AAAI Conference on Artificial Intelligence
  \textbf{30}(1) (Feb 2016),
  \url{https://ojs.aaai.org/index.php/AAAI/article/view/10293}

\bibitem{GaoWei}
Gao, W., Zhang, T., Yang, B.B., Zhou, Z.H.: On the noise estimation statistics.
  Artificial Intelligence  \textbf{293},  103451 (2021).
  \doi{10.1016/j.artint.2021.103451}

\bibitem{GH03}
Grasedyck, L., Hackbusch, W.: Construction and arithmetics of
  {$\mathcal{H}$}-matrices. Computing  \textbf{70}(4),  295--334 (2003)

\bibitem{Part1}
Hackbusch, W.: A sparse matrix arithmetic based on {$\mathcal{H}$}-matrices.
  {I}. {I}ntroduction to {$\mathcal{H}$}-matrices. Computing  \textbf{62}(2),
  89--108 (1999)

\bibitem{HackHMEng}
Hackbusch, W.: Hierarchical matrices: Algorithms and Analysis, Springer Series
  in Comp. Math., vol.~49. Springer (2015)

\bibitem{Huang}
Huang, K., Shi, Y., Zhao, F., Zhang, Z., Tu, S.: Multiple instance deep
  learning for weakly-supervised visual object tracking. Signal Processing:
  Image Communication  \textbf{84},  115807 (2020).
  \doi{10.1016/j.image.2020.115807}

\bibitem{kaya2019predicting}
Kaya, H., T{\"u}fekci, P., Uzun, E.: Predicting co and nox emissions from gas
  turbines: novel data and a benchmark pems. Turkish Journal of Electrical
  Engineering \& Computer Sciences  \textbf{27}(6),  4783--4796 (2019)

\bibitem{khoromskij2009application}
Khoromskij, B.N., Litvinenko, A., Matthies, H.G.: Application of hierarchical
  matrices for computing the {K}arhunen--{L}o{\`e}ve expansion. Computing
  \textbf{84}(1-2),  49--67 (2009)

\bibitem{litv17Tensor}
{Litvinenko}, A., {Keyes}, D., {Khoromskaia}, V., {Khoromskij}, B.N.,
  {Matthies}, H.G.: {Tucker Tensor analysis of Matern functions in spatial
  statistics}. Computational Methods in Applied Mathematics  (Nov 2018).
  \doi{https://doi.org/10.1515/cmam-2018-0022}

\bibitem{litvHLIBPro17}
Litvinenko, A., Kriemann, R., Genton, M.G., Sun, Y., Keyes, D.E.: {HLIBCov}:
  Parallel hierarchical matrix approximation of large covariance matrices and
  likelihoods with applications in parameter identification. MethodsX
  \textbf{7},  100600 (2020). \doi{10.1016/j.mex.2019.07.001},
  \url{https://github.com/litvinen/HLIBCov.git}

\bibitem{LitvGentonSunKeyes17}
Litvinenko, A., Sun, Y., Genton, M.G., Keyes, D.E.: Likelihood approximation
  with hierarchical matrices for large spatial datasets. Computational
  Statistics $\&$ Data Analysis  \textbf{137},  115--132 (2019).
  \doi{10.1016/j.csda.2019.02.002},
  \url{https://github.com/litvinen/large_random_fields.git}

\bibitem{Matern1986a}
Mat\'{e}rn, B.: Spatial Variation, Lecture Notes in Statistics, vol.~36.
  Springer-Verlag, Berlin; New York, second edition edn. (1986)

\bibitem{Muhlenbach}
Muhlenbach, F., Lallich, S., Zighed, D.A.: Identifying and handling mislabelled
  instances. Journal of Intelligent Information Systems  \textbf{22}(1),
  89--109 (2004). \doi{10.1023/A:1025832930864}

\bibitem{Raykar}
Raykar, V.C., Yu, S., Zhao, L.H., Valadez, G.H., Florin, C., Bogoni, L., Moy,
  L.: Learning from crowds. Journal of Machine Learning Research
  \textbf{11}(43),  1297--1322 (2010),
  \url{http://jmlr.org/papers/v11/raykar10a.html}

\bibitem{van2020survey}
Van~Engelen, J.E., Hoos, H.H.: A survey on semi-supervised learning. Machine
  Learning  \textbf{109}(2),  373--440 (2020)

\bibitem{Xiao}
Xiao, Y., Yin, Z., Liu, B.: A similarity-based two-view multiple instance
  learning method for classification. Knowledge-Based Systems
  \textbf{201-202},  105661 (2020). \doi{10.1016/j.knosys.2020.105661}

\bibitem{ZhouD}
Zhou, D., Bousquet, O., Lal, T.N., Weston, J., Sch\"{o}lkopf, B.: Learning with
  local and global consistency. In: Proceedings of the 16th International
  Conference on Neural Information Processing Systems. p. 321–328. NIPS'03,
  MIT Press, Cambridge, MA, USA (2003)

\bibitem{ZhouZH}
Zhou, Z.H.: Ensemble Methods: Foundations and Algorithms. Boca Raton: CRCPress
  (2012)

\bibitem{Zhou}
Zhou, Z.H.: {A brief introduction to weakly supervised learning}. National
  Science Review  \textbf{5}(1),  44--53 (08 2017). \doi{10.1093/nsr/nwx106},
  \url{https://academic.oup.com/nsr/article-pdf/5/1/44/31567770/nwx106.pdf}

\end{thebibliography}

\end{document}